\documentclass[twocolumn,cleanfoot,cleanhead,10pt]{asme2e}
\usepackage{graphicx}
\usepackage{amsmath,amssymb,amsfonts} 
\usepackage{algorithm2e}
\usepackage{color}
\usepackage{changes}
\usepackage{epsfig} 
\usepackage{pstool}
\usepackage{subcaption}

\usepackage{amsthm}
\renewenvironment{proof}{{\bfseries Proof}}{\qed}

\theoremstyle{definition}
\newtheorem{definition}{Definition}

\newtheorem{rem}{Remark}
\newtheorem{problem}{Problem}
\newtheorem{assumption}{Assumption}
\confshortname{DSCC 2018}
\conffullname{ASME 2018 Dynamic Systems and Control Conferences}
\confdate{24-28}
\confdate{September 30-October 3}
\confyear{2018}
\confcity{Atlanta}
\confcountry{USA}

\papernum{DSCC2018-9161}

\title{Human-Robot Trust Integrated Task Allocation and Symbolic Motion planning for Heterogeneous Multi-robot Systems}

\author{Huanfei Zheng
    \affiliation{
	Mechanical Engineering Department\\
	Clemson University\\
	Clemson, South Carolina 29634\\
    Email: huanfez@clemson.edu
    }	
}

\author{Zhanrui Liao 
    \affiliation{Mechanical Engineering Department\\
    Clemson University\\
	Clemson, South Carolina 29634\\
    Email: zhanrul@clemson.edu
    }
}
\author{Yue Wang 
    \affiliation{Mechanical Engineering Department\\
    Clemson University\\
	Clemson, South Carolina 29634\\
    Email: yue6@clemson.edu
    }
    \thanks{This work is partially supported by the Air Force Office of Scientific Research Young Investigator Program under grant no. FA9550-17-1-0050.}
}

\begin{document}

\maketitle    

\begin{abstract}
{\it This paper presents a human-robot trust integrated task allocation and motion planning framework for multi-robot systems (MRS) in performing a set of tasks concurrently. A set of task specifications in parallel are conjuncted with MRS to synthesize a task allocation automaton. Each transition of the task allocation automaton is associated with the total trust value of human in corresponding robots. Here, the human-robot trust model is constructed with a dynamic Bayesian network (DBN) by considering individual robot performance, safety coefficient, human cognitive workload and overall evaluation of task allocation. Hence, a task allocation path with maximum encoded human-robot trust can be searched based on the current trust value of each robot in the task allocation automaton. Symbolic motion planning (SMP) is implemented for each robot after they obtain the sequence of actions. The task allocation path can be intermittently updated with this DBN based trust model. The overall strategy is demonstrated by a simulation with 5 robots and 3 parallel subtask automata.}
\end{abstract}

\section{Introduction}
Symbolic motion planning (SMP) solves complex motion planning problems for robots using linear temporal logic (LTL), languages and automata theory \cite{belta2007symbolic}. It enables the automatic control of a robot or teams of robots from high level with different task specifications. However, computationally efficient frameworks are often needed to deal with the increasing complexity of task specifications and multi-robot system (MRS). Many centralized and decentralized frameworks have been developed to deal with the ``state-space explosion" problems. 
A compositional multi-robot motion planning framework in \cite{saha2014automated} uses precomputed motion primitives for robots and employs a satisfiability modulo theory solver to synthesize robot trajectories. 
Event-based synchronization approach is proposed in \cite{tumova2015decomposition} to address interdependencies among robots in motion planning. 
In \cite{guo2015multi,guo2017task}, a bottom-up strategy is proposed where each robot is assigned with a local task and inter-robot dependence is achieved through cooperative motion and task planning. 
In \cite{ding2011automatic,chen2012formal}, a top-down framework is presented for the automatic deployment of a robotic team from a specification by giving each robot the capabilities to serve the cooperation requirements.
Supervisor synthesis with compositional verification techniques is utilized to guarantee robot performance in \cite{dai2016learning}, where a given team mission is decomposed into individual tasks.

The above multi-robot motion planning frameworks are however restricted either in scalability in terms of the size of robot teams, or in complexity of tasks due to the inter-dependencies among robots. The robot-task pairs are given as fixed in dealing with the global task specifications. That is, these works do not consider the task allocation problem. In this paper, we will first establish a multi-robot multi-task task allocation framework to guarantee the reachability of tasks and optimal assignment of robots. The motion planning of each robots is implemented sequentially based on the task allocation results. Reallocations can also be triggered in this automatic process to deal with the uncertainties of the motion planning in the dynamic environment.  

Under our proposed framework, SMP can therefore ensure automatic and scalable solutions for MRS  motion planning. Moreover, human supervision may facilitate the efficiency and safety of MRS in a dynamic and uncertain environment because humans excel in complex decision-making and robot's performance in such scenarios is not usually not satisfactory.  
Various designs and analyses related with human-robot team cooperation have been conducted to improve human's situation awareness of robot. The work in
\cite{kruijff2014designing} focuses on the development and evaluation of complex socio-technical system for human-robot teaming in Urban Search and Rescue with a user-centric design methodology, which ranges from modeling situation awareness, human robot interaction (HRI), flexible planning, and cognitive system design. 
In \cite{shah2010empirical}, an empirical analysis of human teamwork is conducted to investigate the ways teammates incorporate coordination behaviors, including both verbal and nonverbal cues, into their action planning.
Measurable Shared Mental Models (SMMs) are developed in \cite{nikolaidis2012human} to promote an effective human-robot teaming by observing a team of expert human workers prior to task execution, and then robots executing an interactive planning and cross-training process with a human co-worker to iteratively refine and converge the team model. The framework of discrete-time stochastic hybrid systems is utilized in \cite{vinod2016validation} to model human-in-the-loop cyber-physical systems with discrete choices, and pose the question of expected outcome in terms of a stochastic reachability problem. The paper \cite{chen2018planning} constructs a POMDP based trust model for human to a single robot to improve performance of the joint human-robot system. 

In this work, we will develop a human-robot trust integrated task allocation and motion planning framework for MRS in order to enable a human-like automatic decision-making process for multi-robot tasking. The contribution of the paper is two-fold. First, we synthesize an automatic task (re)allocation framework that can generate solutions with maximum human-robot trust for the system. It enables real-time updating of task allocation of robots in a human-like way. Furthermore, we construct a dynamic Bayesian network (DBN) based human-robot trust model. This model will evaluate the robot performance, safety, human cognitive workload, and the task (re)allocation framework in a system wide trust setting.

The organization of the rest of the paper is as follows. Section \ref{sec:problem} provides the problem setup with a schematic of the human supervised MRS. Section \ref{sec:motionPlanning} describes the human-robot trust associated task allocation framework and symbolic motion planning of each robot for a set of parallel subtasks. Section \ref{sec:trustModel} details the construction of the DBN based human-trust model, and integrates the human-robot trust evaluation into task (re)allocation and motion planning framework. A simulation in Section \ref{sec:simulation} demonstrates the viability and effectiveness of the proposed framework and Section \ref{sec:conclusion} concludes this paper.  

\section{Problem Setup}\label{sec:problem}
\begin{figure*}[t]
    \centering
    \includegraphics[width=5.0in]{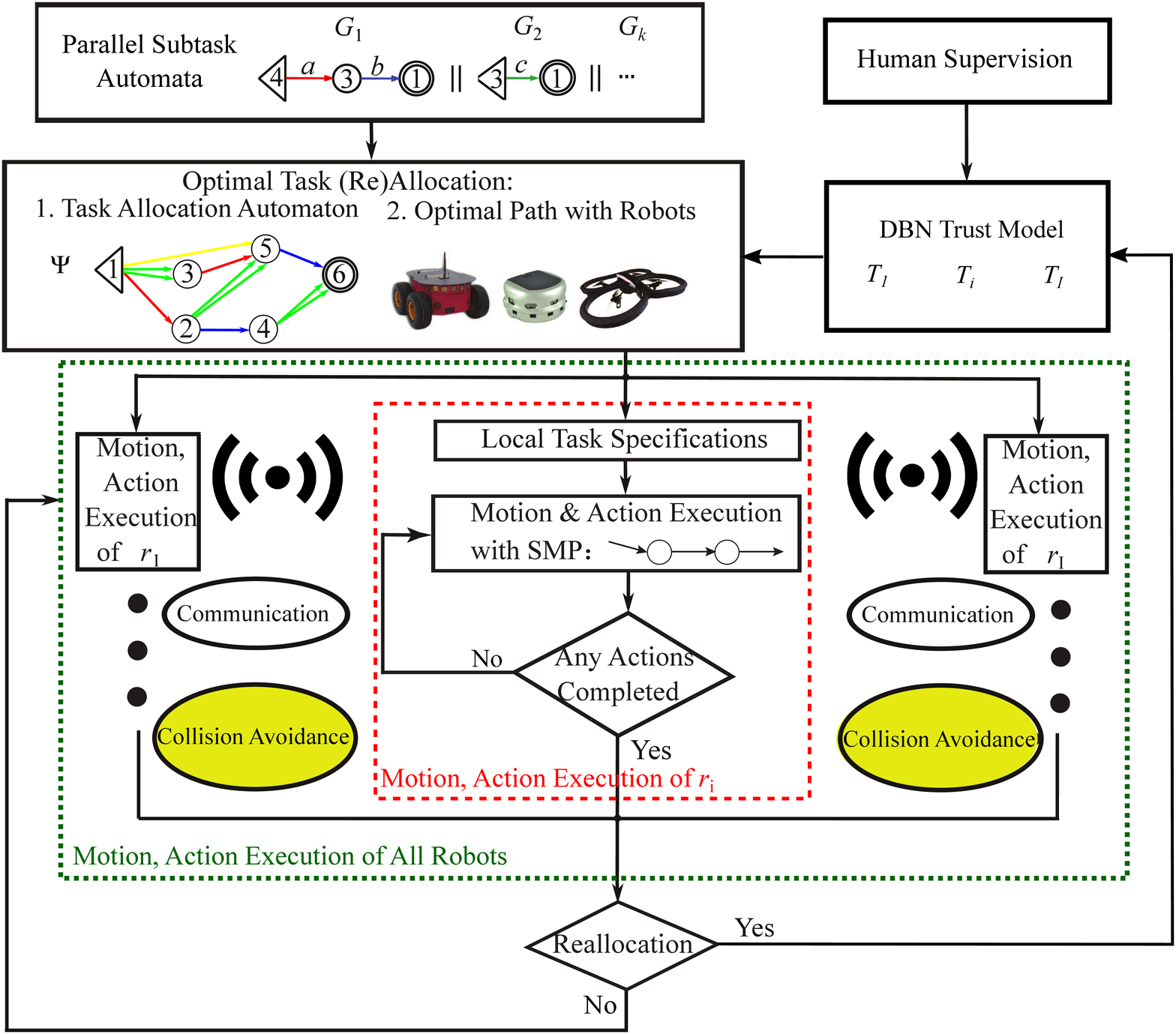}
	\caption{Human-robot trust integrated task allocation and motion planning framework. $\{a,~b\}, ~\{c\}$ are the action sets of parallel subtask automata. $T_1,\cdots,T_i,\cdots, T_I$ are the respective trust value of robots. $\Psi$ is the synthesized task allocation automaton.}
	\label{fig:framework}
\end{figure*}

The schematic of human-robot trust integrated task allocation and SMP is shown in Figure \ref{fig:framework}. Initially, a task allocation automaton is synthesized for a task requirement that concurrently implementable subtasks are to be performed by a set of heterogeneous robots. The subtasks are described by automata and each robot can perform multiple actions in the subtasks. A task allocation path is generated with the maximum accumulated trust of robots from the task allocation automaton. Local action and motion specifications of each robot are mapped from the maximal trust encoded task allocation path so that each robot can execute the motions and actions sequentially. The SMP will also deal with the obstacle collision avoidance in the discrete environment. All these performance will be evaluated to contribute to the calculation of the computational human-robot trust model.

On the other hand, human is allowed to participate into this task allocation and motion planning process in order to improve system performance and reliability. A system-wide human-robot trust model is constructed based on the MRS task allocation by considering robot performance on task performing, safety evaluation on malfunctioning, human cognitive workload of supervision and inter-robot influence from task allocation. The trust of human in each robot will be updated with the progress of task performing. The system-wide trust model will increase or decrease the trust of robots involved with task allocation to construct the interdependence relationship among robots. Once an action is completed by a robot, human will be inquired for the task reallocation based on his/her trust in the current robot. Finally, the parallel subtasks will be completed with a maximum trust encoded task allocation solution and motion planning paths by intermittently updating the task allocation under human supervision.

We summarize the human-robot trust integrated multi-robot task allocation and motion planning with the following assumption and problem.

\begin{assumption}
Each heterogeneous robot $r_i\in \mathcal{R}$ is associated with an action set  $Er_{i}$ describing its capabilities in performing tasks. Assume that $\cup _{i=1}^{I}Er_{i}\supseteq E_{g}$, i.e. all the subtasks can be collaboratively performed by the MRS. Each robot may be able to perform multiple actions, but it is assumed that a robot can only perform one action at a time. 
\end{assumption}

\begin{problem}
\label{prob:problem1}
Given a set of parallel subtasks, described with automata $\{G_k,k=1,\cdots,K\}$, each is associated with an action set $E_k$ that robots need to perform, and all actions are scattered in a dynamic environment, design a task (re)allocation framework such that these subtasks can be completed by $I$ heterogeneous robots $r_i\in \mathcal{R}$ with corresponding capability action set $Er_{i}$, where $i\in \{1,\cdots,I\}$; In the meantime, a human-robot trust model is integrated into the MRS to enable trust-based task reallocation such that human-like decision-making can be deployed for multi-robot multi-task allocation.
\end{problem}


\section{Task Allocation and Symbolic Motion Planning}\label{sec:motionPlanning}

The subtasks of MRS can be described with a set of automata $\{G_k,k=1,\cdots,K\}$. These subtasks satisfy a parallel relationship with $\|_{k=1}^K G_k\triangleq ((G_1\|\cdots G_k)\|\cdots \|G_{K-1})\|G_{K}$, which represent a set of subtask automata $G_k$s that can be concurrently dealt with using a team of robots. We will construct the task allocation automaton with trust associated transitions for these parallel subtasks and robots so that a maximal trust encoded task allocation solution can be generated for the MRS. 

\subsection{Human-Robot Trust Associated Task Allocation}

In order to find a maximum trust encoded task allocation solution, we synthesize a task allocation automaton $\Psi$ by taking into account both robot capabilities and subtask automata. A path from task allocation automaton gives a task allocation solution regarding what actions to be allocated to what robots. To explain the process of generating a maximum trust encoded path from the task allocation automaton, we introduce the following definitions. 


\begin{definition}[\bf Minimal Suffix Set of Language]
\label{def:Pre-Suffix}
Given an automaton $G$ with an action set $E$, the minimal suffix set of language $L(G)$ is denoted as $\mathcal{L}=\{\ell \in E^*: \ell\text{ is the suffix of } \min(s), ~s\in L(G)\}$, where $\min(s)$ is one of the minimal length paths
in $L(G)$ and $E^*$ is the Kleene-closure of $E$.
\end{definition}



\begin{definition}[\bf Implementable Action Set]
\label{def:Impleaction}
Given a task automaton $G$ with an action set $E$, an action $e\in E$ is said to be implementable for a robot $r$ with a capability action set $Er$ if the following two conditions are satisfied: 1) $e\in Er$, which means the action $e$ can be performed by robot $r$, and 2) $e\in \{\ell(0)\}$, i.e. action $e$ is ready to be performed at the states of $G$ that match $\ell$, where $\ell(0)$ is the first element of $\ell$. Hence, an implementable action set of automaton $G$ for robot $r$ is denoted as $IA=\{(r,e) :e\in Er\cap\{\ell(0)\}\}$.
\end{definition}

Accordingly, the implementable action set of automaton $G_k$ for all robots in the set $\mathcal{R}$ can be given as $IG_k=\cup_{i=1}^I IA_{i,k}$, where $IA_{i,k}=\{(r_i,e_{k}) :e_{k}\in Er_{i}\cap\{\ell_k(0)\}\}$. 

\begin{definition}[\bf Multi-action Set]
\label{def:Multi-action}
The set $Act_{\psi}=\{act_\psi\triangleq(\omega_1,\cdots,\omega_k,\cdots,\omega_K):\omega_k=(r_i,\hat e_k), r_i \in \mathcal R, \hat e_k\in Er_{i}\cap\{\ell_k(0)\}\cup \{\epsilon\}\}$ defines a multi-action set, where the single-action $\omega_i=(r_i, \hat e_k)$, $\hat e_k\neq\epsilon$ defines an action in automaton $G_k$ performed by a robot in $r_i\in \mathcal{R}$; $\omega_k=(r_i,\epsilon)$ means no action in $G_k$ is assigned to robot $r_i$. A multi-action $act_{\psi}\in Act_{\psi}$ holds if and only if it is (1) effective, i.e. $\exists \omega_k\not \in \{(r_i,\epsilon),i=1,\cdots,I\}$, (2) unique, i.e. $\forall \omega_k, \omega_{k'} \not \in \{(r_i,\epsilon), i=1,\cdots,I\}$, if $k\neq k'$, then $r_i\neq r_{i'}$. For simplicity of notation, we further denote $\omega_k=(r_i,\epsilon)$ as $\mathcal{E}$. 
\end{definition}

\begin{rem}
The multi-action set combines multiple implementable actions and guarantees not only the state transition of each subtask automaton but also the mutual exclusion of subtask automata for robots.
\end{rem}

Finally, we can define the task allocation automaton for multiple robots to perform multiple tasks as follows.

\begin{definition}[\bf Task Allocation Automaton]
\label{def:Psi}
\sloppy The task allocation automaton $\Psi$ describes the assignment of robots to the actions in subtask automata. It is given by a tuple \mbox{$\Psi= (X_{\psi},Act_{\psi},\delta_{\psi},\chi_{0},W_{\psi})$} where 
\begin{enumerate}
\item $X_{\psi}=\mathcal{L}_1\times \cdots \mathcal{L}_k \times \cdots \mathcal{L}_K$ is the composite state set of task allocation including a set of composite states $\chi=(\ell_{1},\cdots,\ell_{k},\cdots,\ell_{K})$ for the parallel processes $\|_{k=1}^K G_k$, $\ell_{k}\in \mathcal{L}_k$, where $\mathcal L_k$ is the minimal suffix set of $L(G_k)$,
\item the multi-action set $Act_{\psi}\subseteq\prod_{k=1}^K({IG}_{k} \cup\{\mathcal E\})$ (Def.~\ref{def:Multi-action}) 
includes a finite set of actions $act_{\psi}\in Act_{\psi}$ that the heterogeneous robots group $\mathcal{R}$ can perform, 
\item the transition relation $\delta_{\psi}(\chi,act_{\psi})$ is a process $\chi \xrightarrow{act_{\psi}} \chi'$, which can be detailed as $(\ell_1\xrightarrow{\omega_1}\ell_1',\cdots,\ell_K\xrightarrow{\omega_K}\ell_K')$,
\item $\chi_{0}$ is the initial state of task allocation, 
\item $W_{\psi}: Act_{\psi}\rightarrow \mathbb{R}$ is the set of accumulated trust of all robots associated with the completion of each action, $W_{\psi}(act_{\psi})=\sum_{i=1}^I T_i(t)$, where $T_i(t)$ is the trust of human in a single robot $r_i$ at time $t$. 
\end{enumerate}
A finite path $\mathbb S_{\Psi}=act_{\psi}^{(0)}\cdots act_{\psi}^{(\tau)}\cdots act_{\psi}^{(\mathbf T)}$ with $act_{\psi}^{(\tau)}\in Act_{\psi}$ presents a task allocation solution for the parallel processes $\|_{k=1}^K G_k$ with robots $r_i$, $i=1,\cdots, I$.
\end{definition}

An initial task allocation $\mathbb S_{\psi}$ with maximum accumulated trust of all robots $\max(\sum_{\tau=0}^{\mathbf T} W_{\psi})$ from initial state $\chi_{0}$ to final state $\chi_{F}$ can be generated by searching the task allocation automaton $\Psi$. The maximum trust encoded path presents the optimal assignment of robots for all actions in a human-like decision-making pattern, since the associated trust values in task allocation automaton are evaluated with the impact factors in Section \ref{sec:trustModel}, such as robot performance, safety, human cognitive workload, and system wide trust evaluated task allocation. 
The accept state is reached when all subtask automata are reduced to be empty and all its actions are completed. The parallel process based task allocation is conducted among heterogeneous robots without subtask inter-dependency.

\subsection{Symbolic Motion Planning}
The maximum trust encoded task allocation path may provide a task performing sequence for each robot in a human-like decision-making process, but it is also necessary to consider how to deal with the reachability of all these actions in the dynamic environment.
For SMP with a team of robots, paths satisfying the global specification can be generated by model checking, which encodes each robot a single path in the abstracted workspace. 

In this work, we assume that the task environment is not  known a priori, which is also an important prerequisite for robot performance estimation on obstacle avoidance to be discussed in Section \ref{sec:trustModel}. The paths of motion planning for robots are intermittently replanned upon the information they get through exploring the area. 
The mapping of $\mathbb S_{\psi}$ into each $r_i$ from initial step to step $\mathbf{T}$ will give each robot a task allocation path ${s}_{\psi,i}={\omega}_{k_0}^{(0)}\cdots {\omega}_{k_{\tau}}^{(\tau)}\cdots {\omega}_{k_{\mathbf{T}}}^{(\mathbf{T})}$, where $k_{\tau}\in\{1,\cdots,K\}$ is an index of subtask automaton. 

Denote ${s}_{\pi,i}={\pi}_{k_0}^{(0)}\cdots {\pi}_{k_{\tau}}^{(\tau)}\cdots {\pi}_{k_{\mathbf{T}}}^{(\mathbf{T})}$ as the corresponding sequence of motion specifications of $r_i$. ${\pi}_{k_{\tau}}^{(\tau)}$ describes the reachability of ${\omega}_{k_{\tau}}^{(\tau)}$ with ``a robot $r_i$ will go to the position of action $e_k\in E_k$ if and only if it finds that the previous action $\omega_{k_{\tau}}^{(\tau-1)}$ has been completed". Thus, it requires each $\pi_{k_{\tau}}^{(\tau)}$ to be conducted before $\omega_{k_{\tau}}^{(\tau)}$.
The motion specification can guarantee the actions to be conducted in a logic sequence by robots in the decentralized multi-robot motion planning, and every robot obtains the information about completion of action $\omega_{k_{\tau}}^{(\tau-1)}$ by traveling to the location itself.

\begin{definition}[\bf Product Automaton of Robot] \label{def:prodAut}
Given an automaton $A=\left(X, E, f, x_{0}, X_{m}\right)$ and a robot transition system $TS= \left(Q, \delta, q_{init}, \pi, L_{q}, W_{q}\right)$.
We define the product automaton $\mathcal{P} = TS \times A = (\hat{X},E,\hat{f},\hat{x}_{0},\hat{X}_{m}, W_{q})$, where 
\begin{enumerate}
\item $\hat{X} = Q\times X$ is the state set,
\item $E$ is the event set for the transitions,
\item $\hat{f}(\hat{x}, e') = \hat{x}'$ is the transition relation with $\hat{x}=(q,x), ~\hat{x}'=(q',x')$, $q\rightarrow q', ~f(x, e')= x'$, where $q,q' \in Q$ and $x,x' \in X$,
\item $\hat{x}_{0} = (q_{init},x_{0})$ is the initial state, $\hat{X}_{m} = Q \times X_{m}$ is the final state set,
\item $W_{q}$ is the cost set for the transitions in $\delta$. 
\end{enumerate}
\end{definition}

The transition system of robot $r_i$ is abstracted as $TS_{i,k}^{(\tau)}$ for each action $e_k\in E_k$ it is going to perform at step $\tau$ in the discrete space (e.g. see Figure \ref{fig:systemConfig}).
$\mathcal{A}_{k,i}^{(\tau)}$ is an automaton representation of the motion specification $\pi_{k_{\tau}}^{(\tau)}$ of robot $r_i$ regarding subtask automaton $G_k$. The model checking $TS_{i,k}^{(\tau)}\times \mathcal{A}_{k,i}^{(\tau)}$ can provide a motion planning path $\sigma_{i,k}^{(\tau)}$ satisfying a motion specification $\pi_{k_{\tau}}^{(\tau)}$.

Collision avoidance can be dealt within the transition system through a reactive approach. Each robot $r_i$ detects its abstracted surrounding area and stores the detected obstacles in the obstacle set $Obs_{i}^{(t)}$. These robots regard the obstacles as inaccessible states in its transition system $TS_{i,k}^{(\tau)}$.
In addition, neighboring robots in communication ranges are required to exchange the information of respective obstacles and next states. 
The transition system 
is updated after the robot completes the current allocated task or detects new obstacles. 

\section{Human-Robot Trust Model of Task Allocation} \label{sec:trustModel}
\subsection{DBN based Human-Robot Trust Model}
The human-robot trust model is developed to improve the task allocation and motion planning of MRS to be similar as human decision-making. The trust evaluation of human in each robot is involved with robot performance, risk of occurrence of malfunctioning, and human cognitive workload. The robot performance evaluation is dependent on the amount of tasks completed and the success of obstacle collision avoidance. Risks are defined as the occurrence of malfunctioning situations such as robots are unable to move or perform tasks due to low battery level. The cognitive workload is related with the complexity of surrounding environment, such as the amount of surrounded obstacles of each robot, as well as the amount of robots that human has to supervise after a task reallocation. These are all the possible factors that may influence human's interaction with multi-robot task allocation and motion planning. Hence, it is favorable for the MRS to have a human-robot trust model to integrate all these factors in order to enable human-like decision-making for task allocation. 

Besides the above influence from MRS, human and environment, we also consider system-wide trust based influence (either positive or negative) into human-robot trust evaluation regarding task reallocation. That is, robots assigned with an action in the task reallocation will be given an opposite trust evaluation with other robots that have a common implementable action but are not selected for this action. Such influence on MRS will construct a system-wide trust inter-relationship among robots. Finally, we will utilize a DBN based human-robot trust model to assist MRS in task (re)allocation and motion planning. Human will be intermittently inquired whether to allow a task reallocation with this model. 

\begin{figure}
	\centering
    \includegraphics[width=3.2in]{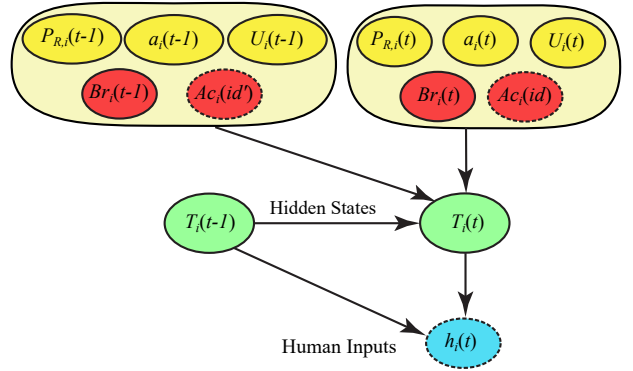}{
}
	\caption{A dynamic Bayesian network (DBN) based model for dynamic, quantitative, and probabilistic trust estimates.}\label{fig:hmm}
\end{figure}

The DBN trust model\footnote{The DBN human-robot model in this paper deals with the human input with respect to each robot individually. In our future work, we will  determine the human input based on the trust of all robots involved in task allocation.} is shown in Figure~\ref{fig:hmm}.
Based on the DBN model, we denote the belief update of trust $T_i(t)$ for robot $r_i$ as 
\begin{align}
\label{eqn:belT}
\begin{split}
bel(T_i(t))=&\textrm{Prob}(T_i(t)|P_{R,i}(1:t),a_i(1:t),U_i(1:t),\\
& Br_{i}(1:t),Ac_i(1:t),h_i(1:t),T_i(0)),
\end{split}
\end{align} 
where $\left[P_{R,i}, a_i, U_i, Br_{i}, Ac_i \right]^T$ are the impacting factors of the hidden trust state, denoted by $\Omega_i$, and $h_i$ is the observed evidence. To be more specific, $P_{R,i}(\cdot)$ is the accumulated performance evaluation of robot $r_i$, $a_i(\cdot)\in[0,1]$ is the safety coefficient of risk evaluation, $U_i(\cdot)$ is the human cognitive workload due to the obstacle-crowded environment, $Br_{i}(\cdot)$ is the human cognitive workload on supervising the MRS (e.g., monitoring multiple robots decided by the task allocation), $Ac_i(\cdot)$ is the extra positive or negative influence on the robots after human accepting of the task (re)allocation. $h_i(\cdot)$ is the human intervention on whether to allow a task reallocation for the MRS. Note that $Br_{i}(\cdot)$, $Ac_i(\cdot)$ and $h_i(\cdot)$ are only intermittently updated when a task reallocation occurs.
\begin{align}
  \begin{split}
    \Omega_i(t)=
         \begin{cases}
           \left[P_{R,i}(t), a_i(t), U_i(t),Br_{i}(t)\right]^T,\text{ if no reallocation}\\
           \left[P_{R,i}(t), a_i(t), U_i(t),Br_{i}(t),Ac_i(id)\right]^T, \textrm{ otherwise}\\
           \end{cases},
    \end{split}
\end{align} 

A forward algorithm is utilized by applying the principle of dynamic programming to avoid incurring exponential computation time due to the increase of $t$. Eqn. (\ref{eqn:belT}) can be calculated as
\begin{eqnarray}\label{eq:BayesUpdates}
&&
bel(T_i(t))=\frac{\int\overline{bel}(T_i(t),T_i(t-1))\textrm{d} T_i(t-1)}{\int \int \overline{bel}(T_i(t),T_i(t-1))\textrm{d} T_i(t-1)\textrm{d} T_i(t)},
\end{eqnarray}
where 
\begin{align}
  \begin{split}
    \overline{bel}( &T_i(t), T_i(t-1)) =\textrm{Prob}(h_i(t)|T_i(t), T_i(t-1))\cdot \\
& \textrm{Prob}(T_i(t)| T_i(t-1), \Omega_i(t), \Omega_i(t-1)) \cdot bel(T_i(t-1)),\label{eq:filteredbelief}
  \end{split}
\end{align} 

To obtain the belief update of each robot trust, $\textrm{Prob}(h_i(t)|T_i(t), T_i(t-1))$ and $\textrm{Prob}(T_i(t)|T_i(t-1),\Omega_i(t),\Omega_i(t-1))$ are respectively calculated with different distribution models as shown in the upcoming paragraphs.

The term $\textrm{Prob}(h_i(t)|T_i(t),T_i(t-1))$ is the conditional probability of human intervention given the current and prior trust, which can follow a similar sigmoid distribution as in~\cite{xu2015optimo}. Therefore, the conditional probability distribution (CPD) of human intervention based on trust can be modeled as follows
\begin{equation}\label{eq:omega}
\begin{split}
\textrm{Prob}(h_i(t)=&1|T_i(t), T_i(t-1))=\frac{1}{1+exp(-\alpha_1 T_i(t)+\alpha_2 T_i(t-1))},
\end{split}
\end{equation}
where $\alpha_1$ and $\alpha_2$ are positive weights and this CPD indicates higher willingness for human to allow a task reallocation when human-robot trust value is higher.

The CPD of human trust in robot $r_i$ at time $t$ can be constructed based on the previous trust value, robot performance, risk coefficient, human cognitive workload, and task allocation evaluation. It is expressed as a Gaussian distribution with mean value $\bar{T}_i(t)$ and variance $\rho_i(t)$,
\begin{align}
  \label{eq:guass_cpd}
  \begin{split}
    \textrm{Prob}(T_i&(t)|T_i(t-1), \Omega_i(t), \Omega_i(t-1))=\mathcal{N}(T_i(t);\bar{T}_i(t),\rho_i(t)),\\
    \bar{T}_i(t) = & A\cdot \bar{T}_i(t-1) + B_1\cdot a_i(t)\cdot P_{R,i}(t) - B_2\cdot a_i(t-1)\cdot P_{R,i}(t- \\
    & 1) + C_1 \cdot U_i(t) - C_2 \cdot U_i(t-1) + D_1\cdot Br_{i}(t) - D_2\cdot Br_{i}(t\\& -1) + E_1\cdot Ac_i(id) - E_2\cdot Ac_i(id'),
  \end{split}
\end{align}
where $\bar{T}_i(t)\in(0,1)$ represents the mean value of human trust in robot $r_i$ at time $t$, and $\rho_i(t)$ reflects the variance in each individual's trust update. Each parameter is evaluated with a function of respective influence factors in task allocation and motion planning. The coefficients $A,B_1,B_2,C_1,C_2,D_1,D_2,E_1,E_2$ are determined by data collected from human subject tests ~\cite{sadrspringer2015}. 

In our scenario, the accumulated performance evaluation $P_{R,i}(t)$ is modeled as a function of rewards on robot for its completion of actions as well as the avoidance of obstacles,
\begin{align}
\label{eqn:perform}
  \begin{split}
    P_{R,i}(t) & = P_{R,i}(t-1) + w(r_i,\hat e_k,t) + \beta_i(t) \cdot w(o^{(t)}_i),\\
  \end{split}
\end{align}
where $P_{R,i}(t-1)$ is the performance of robot at $t-1$, $w(r_i,\hat e_k,t)\in \{0,1\}$ is the reward on robot $r_i$ for completing an action $\hat e_k\in E_k$, $w(o^{(t)}_i)\in \{0,1\}$ is the reward on robot for avoiding a detected obstacle $o^{(t)}_k$ at $t$, $\beta_i(t)$ is the number of obstacles the robot can avoid by re-planning path at $t$.
The safety coefficient $a_i(t)$ is introduced to evaluate the potential of a single robot in completing all the capable actions in $E_{r_i}$. 
Here, the risk of malfunction refers to the possibility of low battery level of the robot, which may constrain it to perform more actions and thus need other robots to substitute it for the remaining uncompleted actions. The safety coefficient $a_i(t)$ is constructed as
\begin{equation}
   a_{i}(t)=
     \begin{cases}
       \label{eqn:safe}
       1, & ~r_i \text{ is in normal state}\\
       \frac{1}{\vert E_{r_i\vert}}, & ~r_i \text{ is in low battery state}
     \end{cases}.
\end{equation}
This implies the system tends to trust the robot to complete all its capable actions if it has enough electric capacity. On the other hand, if the battery level is low, the robot is assigned to at most complete the current allocated action. 

The human cognitive workload is a result of interaction with the complex environment and multiple robots. For the environmental complexity resulted workload, it is constructed as
\begin{equation}
  U_i(t)=1 - \gamma(t)^{S_{o,i}(t)+1},
\end{equation}
where $S_{o,i}(t)$ is the number of obstacles within sensing range of robot, and $\gamma(t)$ is the utilization ratio~\cite{spencer2015slqr, WaHuLiZh-TiiS-18}.
The human cognitive workload resulted from supervising robots always exists but is only updated after a task reallocation is implemented. It is estimated by the amount of robots that human can deal with as well as the actual activated robots in the supervision of a MRS. It is intermittently updated based on the $id$th task reallocation as
\begin{equation}
   \label{eqn:loadAllo}
   Br_{i}(t)=
     \begin{cases}
       1-I_{act}(id)/I_{\max}, &\text{if $r_i$ is activated}\\
       1, &\text{otherwise}
     \end{cases},
\end{equation}
where 
$I_{act}(id)$ is the actual amount of activated robots in this task reallocation, and $I_{\max}$ is the maximal number of robots that human feels comfortable in supervising the MRS. 
Each robot will be updated the same workload with the system-wide trust theory if it is activated in the task (re)allocation, while $Br_{i}(t) = 1$ if the robot is not activated at all.

The extra positive or negative influence of task reallocation on each robot also works according to the system-wide trust theory after human accepts a task reallocation solution. Recall that each robot has a task reallocation path ${s}_{\psi,i}={\omega}_{k_{0}}^{(0)}\cdots {\omega}_{k_{\tau}}^{(\tau)}\cdots {\omega}_{k_{\mathbf{T}}}^{(\mathbf{T})}$. Opposite influences can be enforced on the following situation: (1) the action in path ${s}_{\psi,i}$ is an implementable action of $r_i$, and (2) the implementable action of $r_i$ is not selected in the current task allocation. As a result, an extra influence of $id$th task reallocation on each robot is constructed as
\begin{align}
\begin{split}
  Ac_{i}(id)=&\sum_{\tau=0}^{\mathbf{T}}ac_{i}(\tau) ,\\
   ac_{i}(\tau)=&
     \begin{cases}
       \label{eqn:subjTrust}
       \mu/I, &\text{if $ {\omega}_{k_{\tau}}^{(\tau)} = IA_{i,k}(\tau)$}\\
       \overline{\mu}/I, &\text{if ${\omega}_{k_{\tau}}^{(\tau)} \neq IA_{i,k}(\tau)$}
     \end{cases},
\end{split}
\end{align}
where $ac_{i}(\tau)$ is the positive or negative influence of each action in task reallocation path on robot $r_i$, $\mu>0$ and $\overline{\mu}<0$ are the influence coefficients. If the action in task reallocation path is an implementable action of $r_i$, a positive influence will be added for this robot, which implies a trust increase of this robot in the current task reallocation; A negative influence will be associated to a robot by decreasing the trust of the robot if the implementable action of the robot is not selected in the current task allocation.

\begin{rem}
The network parameters for the DBN such as $\alpha_1$, $\alpha_2$ in Equation (\ref{eq:omega}) can be learned by the well-known expectation maximization (EM) algorithm~\cite{moon1996expectation} off-line during the training session and hence will not affect the functionality of the system and the user experience in real-time operation. Besides, a separate and personalized trust model should be trained based on each user's experience since the model strongly depends on individual human intervention $h_i(\cdot)$ as well as the impacting factors $\left[P_{R,i}(\cdot), a_i(\cdot), U_i(\cdot), E_{H,i}(\cdot), Ac_i(\cdot)\right]$. 
\end{rem}
\subsection{Human-Robot Trust based Real-time Interactive Task Reallocation and Motion Re-planning}\label{sec:integrated}

Human trust in robot can be updated at each time step $t$ or intermittently after a task reallocation of MRS. Consequently, the previous maximum trust encoded task allocation and motion planning solution need to be updated. 

The reallocation request is triggered after an action is completed by a robot in the MRS and the human-robot trust is higher enough. Human works as a supervisor and will be inquired if he/she would like to have a task reallocation. The system will reallocate the actions to robots and re-plan the motion path of individual robot if human allows to have a task reallocation. The task reallocation will be implemented on these uncompleted actions, i.e. a task allocation automaton is re-synthesized with the unperformed actions constituted state set. As a result, a new maximum trust encoded path is generated from this automaton for the remaining task. In the mean time, the human trust in each activated robot will be changed with $Ac_i(id)$ from a system-wide trust perspective. However, if human refuses the task reallocation, the MRS will continue the previous task allocation and motion planning path. 
On the other hand, the human-robot trust model will be continuously updated for robot performance, safety coefficient, and human cognitive workload estimations while the robot is exploring in the work space. 
Algorithm $1$ describes the complete process of human-robot trust model based interactive task allocation and motion planning. The process is iterated until all actions are completed.

\begin{algorithm}
 \caption{Human-robot trust integrated task (re)allocation and SMP}
  \label{alg:reallocation}
  Initial task allocation $\mathbb S_{\psi,0}$\;
  Update influence $Ac_i(0)$, trust $T_i(t)$ for $r_i \in \mathcal I$\;
  \While{Exist unperformed actions}{
     \If{An action completed}{
     Update $P_{R,i}(t), a_i(t), U_i(t), Br_i(t), T_i(t)$\;
          \If{Allow reallocation} {
          Task reallocation $\mathbb S_{\psi,\tau}$\;
          Update $Ac_i(id),T_i(t)$\;
          Motion planning $\sigma_i$ for all $r_i$\;
     }
     }
     Execute motions and actions\;
     Update $P_{R,i}(t), a_i(t), U_i(t), T_{i}(t)$\;
}
\end{algorithm}

\section{Simulation}\label{sec:simulation}
\subsection{System Configuration and Task Specification}

\begin{figure}[t]
    \centering
    \includegraphics[width = 2.0in]{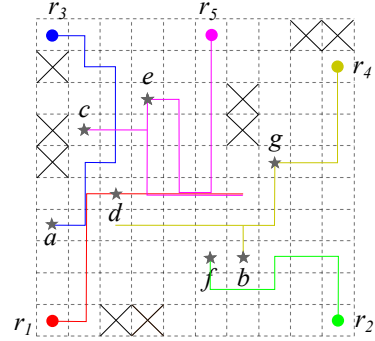}
	\caption{Abstracted workspace and paths of motion planning where ``$\boxtimes$'' marks the obstacle and ``$\bigstar$" marks the allocated task for each robot.}
	\label{fig:systemConfig}
\end{figure}

The workspace of MRS is abstracted as a $10\times 10$ grid environment occupied with obstacles and task stations as shown in Figure~\ref{fig:systemConfig}. Each task station is associated with an action that needs a robot to perform. In the SMP, we assume the motion primitives of each robot are the abstracted from one grid to its adjacent four grids (north, east, south and west). A robot is also assumed to be unable to enter into grids that are partially or totally taken by obstacles. The linear quadratic regulator (LQR) is utilized to control the motion of robots between two grids. 

\begin{figure}[t]
    \centering
    \includegraphics[width = 3.0in]{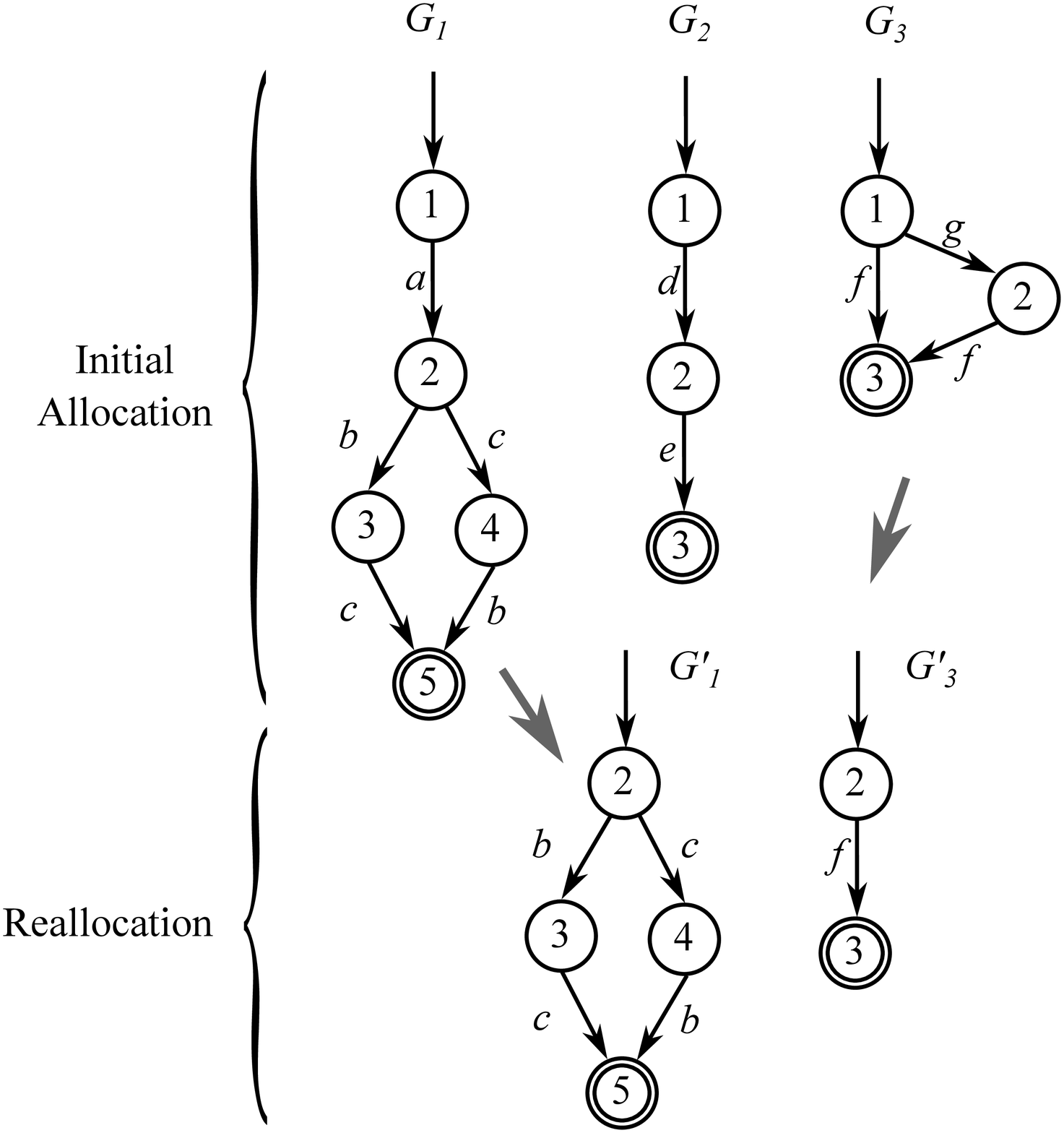}
	\caption{Parallel subtask automata $G_1$, $G_2$ and $G_3$.}
	\label{fig:subtask}
\end{figure}

3 parallel subtasks are to be allocated for MRS, each of the subtask is described by an automaton, see Figure~\ref{fig:subtask}. The languages of the 3 subtask automata are $L(G_1)=\{abc, acb\}$, $L(G_2)=\{de\}$, and $L(G_3)=\{f, gf\}$. 
A team of 5 robots $r_i,~i=1,\cdots,5$ are assigned to perform the actions in the 3 parallel subtasks. Each robot is associated with its capability: $Er_{1}=\{a, c, d\}$, $Er_{2}=\{b, e, f\}$, $Er_{3}=\{a, f, g\}$, $Er_{4}=\{b, d, g\}$, and $Er_{5}=\{c, e\}$. 
In addition, we assume omni-directional sensors and set as two-grid length. The communication radius is set as the same length. 


\subsection{Results}
\begin{figure}[t]
    \begin{subfigure}[b]{0.23\textwidth}
        \centering
        \includegraphics[height=0.8in]{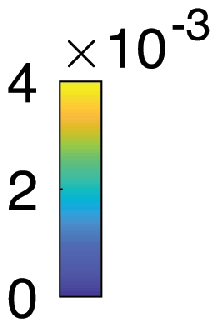}
        \caption{Color bar of trust distribution}
    \end{subfigure}
    \begin{subfigure}[b]{0.23\textwidth}
        \centering
        \includegraphics[width=\textwidth]{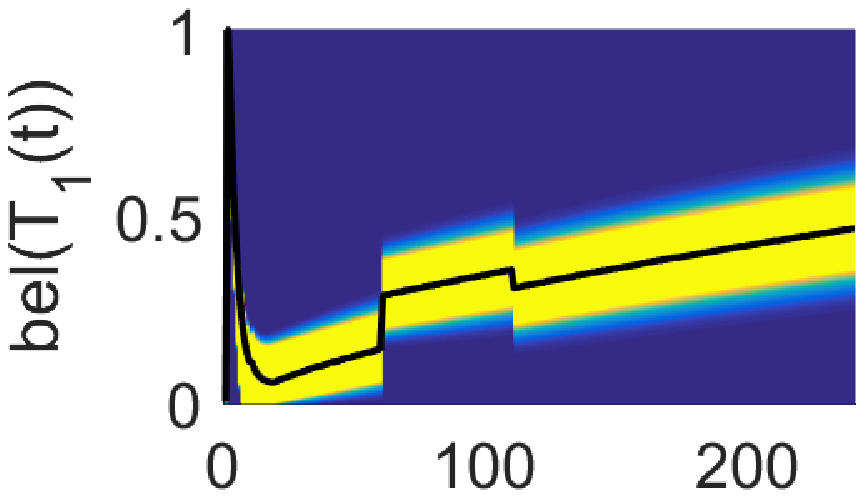}
        \caption{Trust distribution of $r_1$}
    \end{subfigure}
    \begin{subfigure}[b]{0.23\textwidth}
        \centering
        \includegraphics[width=\textwidth]{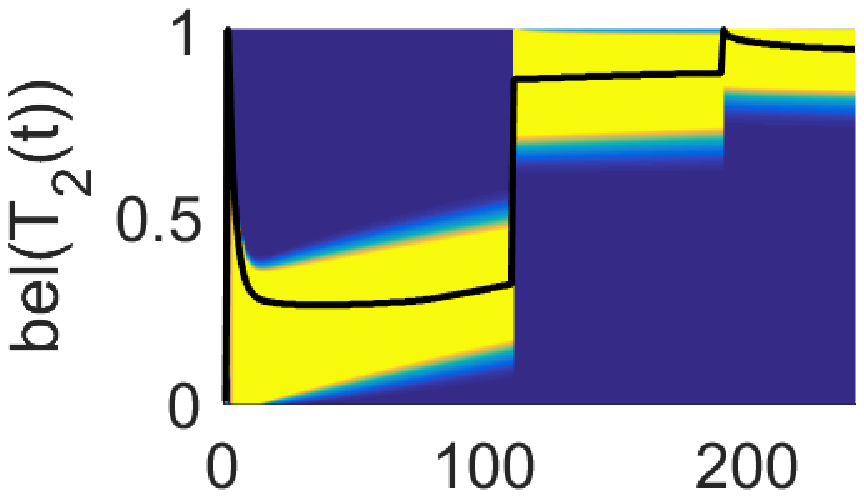}
        \caption{Trust distribution of $r_2$}
    \end{subfigure}
    \begin{subfigure}[b]{0.23\textwidth}
        \centering
        \includegraphics[width=\textwidth]{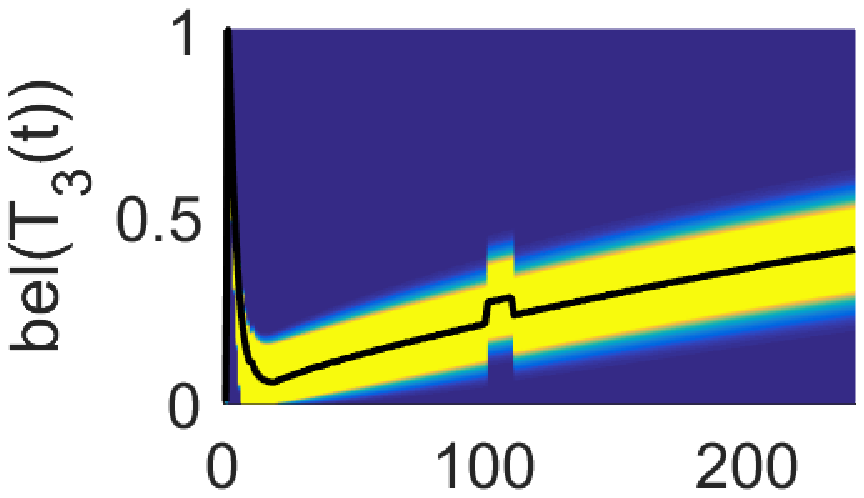}
        \caption{Trust distribution of $r_3$}
    \end{subfigure}
    \begin{subfigure}[b]{0.23\textwidth}
        \centering
        \includegraphics[width=\textwidth]{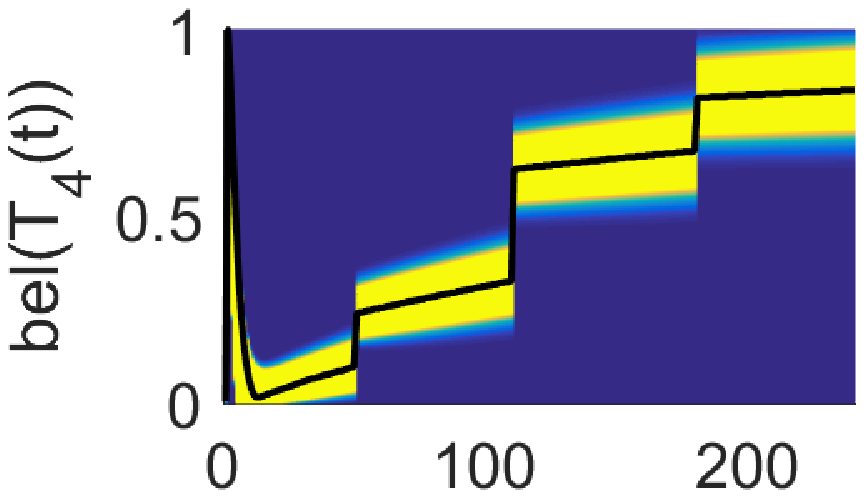}
        \caption{Trust distribution of $r_4$}
    \end{subfigure}
    \begin{subfigure}[b]{0.24\textwidth}
        \centering
        \includegraphics[width=\textwidth]{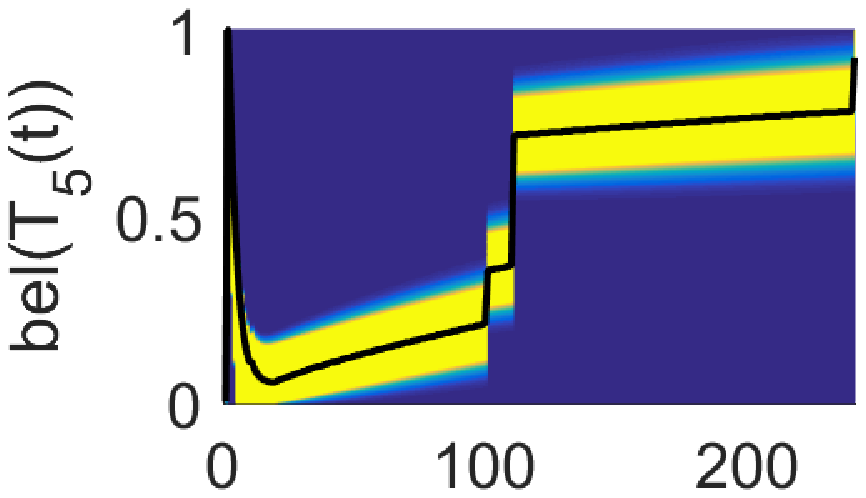}
        \caption{Trust distribution of $r_5$}
    \end{subfigure}
    \caption{Evolution of trust distribution of each robot. Black curves are the maximum values of trust distribution. }\label{fig:trust}
\end{figure}

The final motion paths of robots are shown in Figure \ref{fig:systemConfig}. The corresponding trust change of each robot is shown in Figure \ref{fig:trust}. 
The initial generated task allocation path is $\mathbb S_{\psi}=((r_3,a),(r_1,d),(r_4,g))^{(0)}((r_4,b),(r_5,e),(r_3,f))^{(1)}((r_1,c),\mathcal{E},\mathcal{E})^{(2)}$. The task allocation mappings into each robot are $s_{\psi,1}=(r_1,d)^{(0)}(r_1,c)^{(2)}$, $s_{\psi,3}=(r_3,a)^{(0)}(r_3,f)^{(1)}$, $s_{\psi,4}=(r_4,g)^{(0)}(r_4,b)^{(1)}$ and $s_{\psi,5}=(r_5,e)^{(1)}$. As a result, the trust of each robot will be updated regarding the positive or negative influence of task allocation on each robot. 

Actions $(r_4,g),~(r_1,d),~(r_5,e),~(r_3,a)$ are first sequentially completed with reference to the current task allocation solution and motion specification. Each robot verifies the completion state of actions that need to be performed before their current allocated actions in the subtask automaton. The results are demonstrated in Figure~\ref{fig:systemConfig}, where robot $r_4$, $r_5$ and $r_3$ first go to the neighboring positions of a, d, and g respectively (i.e. within the robot's sensing range) to detect the completion states before they perform the current allocated actions. The action $a$ is almost completed by robot $r_3$ at the same time with $(r_5,e)$. The rewards in performance evaluation of each robot are updated immediately after they completed each assigned actions, and the cognitive workload is also updated during the robot exploration. A reallocation inquiry is triggered after robot $r_5$ completes its only assigned action $e$. The reallocation is synthesized for the remaining subtasks $L(G_1)=\{c\}$ and $L(G_3)=\{f\}$, while action $b$ is still performed by $r_4$ considering the robot's previous effort. The reallocation moment is at $110$ time step, and trust change of each robot is demonstrated in the trust distribution in Figure \ref{fig:trust}. The maximum trust value of each robot before the reallocation are shown in Table \ref{trust_reallocation}. 

\begin{table}[t]
\caption{Maximum trust of each robot before task reallocation.}
\begin{center}
\label{trust_reallocation}
\begin{tabular}{l c c c c c}
& & \\ 
\hline
robot & 1 & 2 & 3 & 4 & 5 \\
\hline
Max trust & 0.3566 & 0.3167 & 0.2818 & 0.3267 & 0.3666 \\
\hline
\end{tabular}
\end{center}
\end{table}
The robots that enable the task allocation associated with the highest accumulated trust are selected to perform the remaining actions. According to the maximum trust value table above, robot $r_2$ and $r_5$ are selected respectively to perform action $f$ and $c$ rather than $r_3$ and $r_1$. The updated task allocation path is $\mathbb S'_{\psi}=((r_3,a)(r_1,d)(r_4,g))^{(0)}((r_4,b),(r_5,e),(r_2,f))^{(1)}((r_5,c),\mathcal{E},\mathcal{E})^{(2)}$. The newly assigned task allocation mapping into each robot are $s_{\psi,2}=(r_2,f)^{(1)}$, $s_{\psi,4}=(r_4,b)^{(1)}$ and $s_{\psi,5}=(r_5,c)^{(2)}$. Since human accepted the reallocation, the positive or negative trust influence on each robot get updated. Eventually, the MRS completes the remaining actions with this updated solution.

\section{Conclusion}\label{sec:conclusion}
This paper presents a human supervised task allocation and motion planning framework for MRS to perform multiple parallel subtasks in a human-like decision making manner. These subtasks are described by automata and conjuncted with MRS to synthesize a task allocation automaton. Transitions of task allocation automaton are associated with the estimations of robot performance and human cognitive workload. They are combined with a DBN human-robot trust model and a maximal trust encoded task allocation path can be found. This path reflects the maximum trust of human in task assignment of MRS. Symbolic motion planning (SMP) is implemented for each robot after the task allocation. The task reallocation is triggered after an action being completed with human permission. The above process is demonstrated to be effective for MRS task allocation by a simulation with 5 robots and 3 parallel subtasks.
\bibliographystyle{asmems4}

%
% ...

\bibliography{asme2e}


\end{document}